\documentclass[10pt,twocolumn,letterpaper]{article}

\usepackage{cvpr}              
\usepackage{mathrsfs}
\usepackage{multirow}
\usepackage{multicol}
\usepackage{tabularx}
\usepackage{graphicx}
\usepackage{algorithm}
\usepackage{algorithmic}
\usepackage{pifont}
\usepackage{hhline} 

\definecolor{cvprblue}{rgb}{0.21,0.49,0.74}
\usepackage[pagebackref,breaklinks,colorlinks,allcolors=cvprblue]{hyperref}

\def\paperID{9355} 
\def\confName{CVPR}
\def\confYear{2025}

\title{Efficient ANN-Guided Distillation: Aligning Rate-based Features of Spiking Neural Networks through Hybrid Block-wise Replacement}

\author{
    Shu Yang$^{1,\dagger}$, Chengting Yu$^{1,2,\dagger}$, Lei Liu$^{1}$, Hanzhi Ma$^{1,2,*}$, Aili Wang$^{1,2,*}$, Erping Li$^{1,2,*}$\\
    $^{1}$ ZJU-UIUC Institute, Zhejiang University\\ 
    $^{2}$ College of Information Science and Electronic Engineering, Zhejiang University 
\thanks{$^{\dagger} $Equal contribution, $^{* }$Corresponding authors: mahanzhi@zju.edu.cn, ailiwang@intl.zju.edu.cn, liep@zju.edu.cn.}
}

\makeatletter
\def\thanks#1{\protected@xdef\@thanks{\@thanks
        \protect\footnotetext{#1}}}
\makeatother

\begin{document}

\maketitle

\begin{abstract}
Spiking Neural Networks (SNNs) have garnered considerable attention as a potential alternative to Artificial Neural Networks (ANNs). Recent studies have highlighted SNNs' potential on large-scale datasets. For SNN training, two main approaches exist: direct training and ANN-to-SNN (ANN2SNN) conversion. To fully leverage existing ANN models in guiding SNN learning, either direct ANN-to-SNN conversion or ANN-SNN distillation training can be employed. In this paper, we propose an ANN-SNN distillation framework from the ANN-to-SNN perspective, designed with a block-wise replacement strategy for ANN-guided learning. By generating intermediate hybrid models that progressively align SNN feature spaces to those of ANN through rate-based features, our framework naturally incorporates rate-based backpropagation as a training method. Our approach achieves results comparable to or better than state-of-the-art SNN distillation methods, showing both training and learning efficiency.

\end{abstract}

\section{Introduction}
\label{sec:1}

Spiking Neural Networks (SNNs) are third-generation neural networks with biological interpretability \cite{maass1997networks,roy2019towards}. Unlike  Artificial Neural Networks (ANNs), SNN neurons simulate the activity of biological neurons, providing both spatial and temporal expressiveness, while outputting binary spikes \cite{panzeri2001unified}. This makes SNNs highly energy-efficient for deployment on neuromorphic hardware \cite{pei2019towards,davies2018loihi,akopyan2015truenorth}, offering promising development prospects.

Current SNN training methodologies primarily follow two paradigms: direct training through Backpropagation Through Time (BPTT) and ANN-to-SNN (ANN2SNN) conversion.
BPTT-based approaches utilize surrogate gradients to approximate the non-differentiable spike function \cite{wu2018spatio,neftci2019surrogate}. While these methods achieve competitive performance with limited timesteps \cite{neftci2019surrogate,shrestha2018slayer,wu2018spatio,li2021differentiable,zhang2020temporal,kim2020unifying,xiao2021training,deng2023surrogate}, they typically encounter increased computational costs due to temporal gradient calculations \cite{xiao2022online,dsr,zhu2024online,yu2024advancing}. Additionally, the surrogate gradients do not always align with the theoretical gradients, requiring further refinement of BPTT-based SNN training.
ANN-to-SNN conversion methods establish analytical mappings between Integrate-and-Fire (IF) neuron firing rates and ReLU activations \cite{rueckauer2017conversion}. Recent approaches adopt an ANN$\rightarrow$QNN$\rightarrow$SNN pipeline \cite{li2022quantization,bu2023optimal,bu2022optimized,jiang2023unified,hu2023fast}, incorporating quantization-aware training (QAT) \cite{bhalgat2020lsq+,yu2024improving} as an intermediate step. However, this introduces mapping errors during QNN-to-SNN conversion, particularly pronounced with reduced timesteps. Furthermore, while empirical evidence suggests Leaky Integrate-and-Fire (LIF)-based architectures outperform IF-based SNNs \cite{hunsberger2015spiking,dsr}, the absence of theoretical support for converting LIF neurons remains a limitation.

Notably, given the inherent challenges of both mainstream training approaches, recent work has drawn inspiration from both ends, proposing innovative approaches that offer unique advantages and insights.
(1) ANN-guided distillation aims to incorporate "dark knowledge" from ANNs into direct SNN training through a non-strict mapping \cite{guo2023joint,qiu2024self,xu2023constructing,zuo2024self,zhangsupersnn}. In this method, the ANN provides supervisory signals, such as logits or feature alignment, which encourages the SNN to align its features with those of the ANN. This creates an implicit ANN-to-SNN conversion \cite{bu2024training}, where the ANN serves as a form of regularization, subtly guiding the SNN's learning.
(2) Recent studies have noted that SNN features are often represented in a rate-based manner \cite{el2021securing,kundu2021hire,sharmin2020inherent}, similar to the feature representations in ANNs.  The similarity in rate-based representations has been further highlighted by recent findings \cite{li2023uncovering}. This has prompted the development of lightweight training methods that optimize SNNs to leverage this alignment with ANN representations \cite{yu2024advancing}.
However, both approaches face limitations. ANN-guided distillation struggles to explicitly align intermediate feature spaces due to the limited expressive power of binary spike representations, hindering effective SNN-ANN feature map matching \cite{qiu2024self,zhangsupersnn}. On the other hand, rate-based backpropagation, though capturing rate-based representations well, is entirely self-driven within the SNN. This self-driven training introduces gradient distortions in intermediate variables due to the Straight-Through Estimator (STE) \cite{bengio2013estimating}, which impedes optimal convergence of rate-based features.

These observations inspire us to explore the integration of ANN-guided training with rate-based representation. While spike-based representations in ANN-guided training capture features over $T$ timesteps, they impose strict alignment constraints with intermediate ANN features \cite{qiu2024self,zhangsupersnn}, underscoring the need for a more flexible alignment approach. By adopting a rate-based backpropagation framework, we leverage a more intuitive and efficient training alternative compared to BPTT’s spike focus. Moreover, integrating an ANN-guided alignment mechanism within rate-based backpropagation may alleviate gradient distortions from the STE, adding a subtle regularization term that guides convergence toward optimal points.

Building on these insights, we delve deeper into the synergy between ANN-guided training and rate-based representations, aiming to bridge the gap between the two. To effectively learn rate-based representations, rather than strictly aligning the intermediate features of ANNs and SNNs, we propose an innovative approach that aligns the mapping relationships between ANN and SNN modules. In other words, instead of hard-aligning features, we encourage the SNN’s input-output mappings to approximate those of the ANN, forming an implicit functional alignment.
Following this idea, we designed an ANN-guided scheme based on block-wise replacement within the rate-based backpropagation framework. In this setup, the terminal modules of the ANN act as auxiliary branches in the feedforward process, allowing the ANN to more effectively guide the learning of rate-based representations. This approach achieves two main goals: (1) flexible alignment of intermediate features between the ANN and SNN, and (2) mitigation of gradient distortion in rate-based backpropagation.
Empirical results demonstrate that our method enables more effective knowledge transfer, ultimately enhancing the SNN’s learning process and achieving state-of-the-art performance across CIFAR-10, CIFAR-100, DVS-CIFAR10, and ImageNet.

Our contributions can be summarized as follows:
\begin{enumerate}
\item  We introduce a framework for ANN-guided learning in SNNs that effectively leverages pretrained ANN knowledge for SNN distillation.
\item By using rate-based representation to construct the guidance signal, our approach enables efficient distillation training using rate-based backpropagation.
\item Our method achieves or surpasses state-of-the-art performance on the ImageNet, CIFAR-10, CIFAR-100, and CIFAR-10-DVS datasets.
\end{enumerate}

\begin{figure*}[t]
\centering
\includegraphics[width=1.0\textwidth]{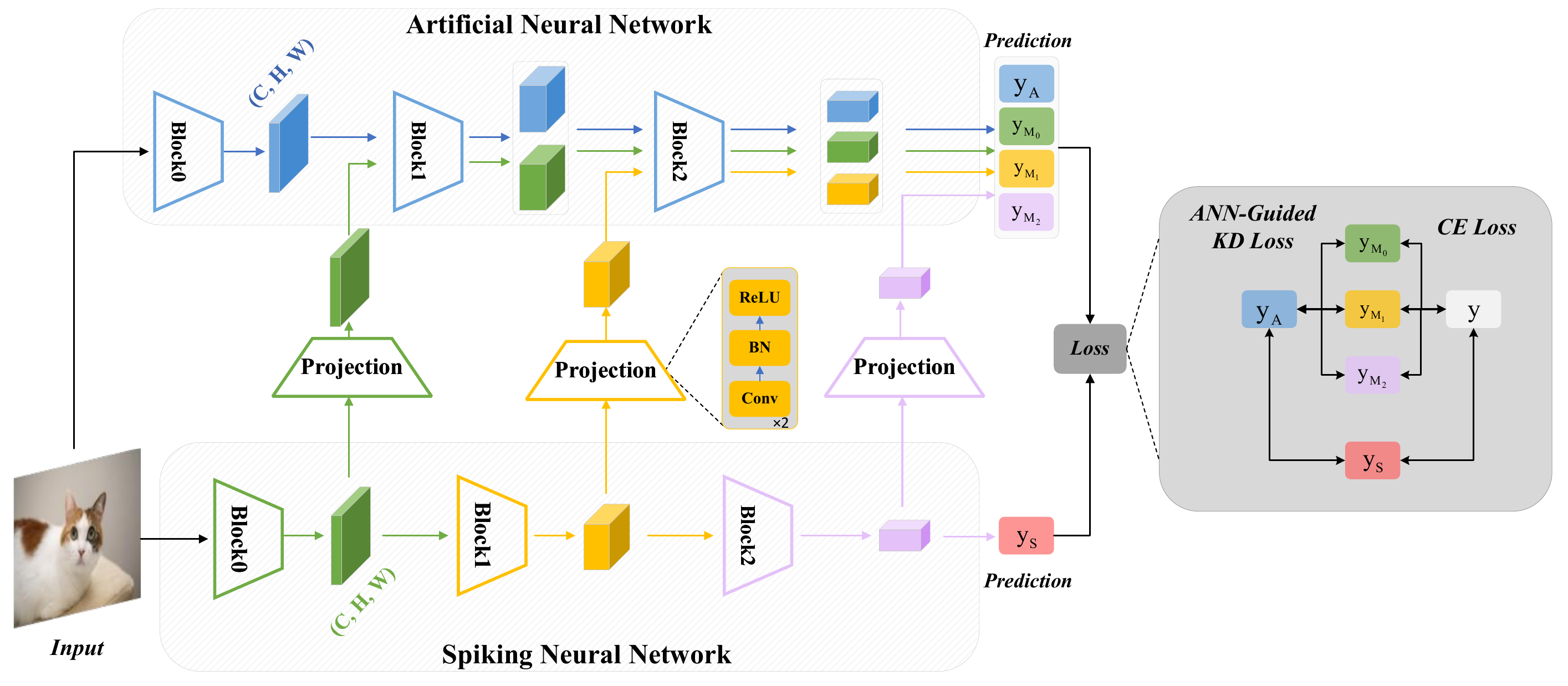} 
\caption{Framework Overview. In the rate propagation of the SNN, the rate-based representation is transformed via learnable modules to align with the ANN’s intermediate feature space and then propagated forward along the subsequent ANN layers, yielding a series of hybrid model outputs, \(\{y_{M_0}, y_{M_1}, y_{M_2}\}\). At the logits end, losses are computed among the outputs of the ANN (\(y_{A}\)), the SNN (\(y_{S}\)), and the hybrid models. The ANN-guided distillation loss interacts with the ANN output, while the standard cross-entropy loss interacts with the hard labels \(y\).}
\label{fig:main}
\end{figure*}

\section{Related Work}
\label{sec:2}

\noindent
\textbf{SNNs Learning Methods} Among primary learning approaches for SNNs, direct training based on BPTT faces significant computational overhead, with memory and time complexity increasing linearly with the number of timesteps \cite{su2024snn,deng2023surrogate,kim2020unifying,li2021differentiable,dsr,xiao2022online,xiao2021training,zhang2020temporal}. Some studies have drawn inspiration from the online learning strategies used in Recurrent Neural Networks (RNNs) \cite{williams1989learning}, proposing online learning methods that retain the distinctive characteristics of SNN neurons, aimed at reducing BPTT’s substantial memory requirements and demonstrated effectiveness on large-scale datasets \cite{dsr,xiao2022online,zhu2024online}. Recently, \cite{yu2024advancing} advanced an efficient direct training method for SNNs by using rate-based representation and backpropagation to decouple memory and training time from timesteps. We aim for the SNN model to learn rate-based representations, and since rate-based backpropagation inherently suffers from gradient estimation errors in intermediate variables, incorporating ANN-guided rate-based representation learning naturally complements this issue.

\noindent
\textbf{ANN-guided Training} Training-required ANN2SNN conversions \cite{li2022quantization,bu2023optimal,bu2022optimized,jiang2023unified,hu2023fast} typically grounded in the explicit mapping theories of QNN to SNNs, first applies QAT \cite{bhalgat2020lsq+,yu2024improving} to obtain a QNN, followed by mapping quantized parameters to SNN neuron parameters. However, these conversion methods remain tied to IF neurons, which limits their ability to leverage the higher-performance LIF neurons. 
Knowledge distillation (KD) is a well-established transfer learning technique effectively utilized for model compression \cite{hinton2015distilling,liu2019structured,sun2019patient,wang2021knowledge,yu2024decoupling}. Recent works have adapted KD for SNNs \cite{kushawaha2021distilling,lee2021energy,takuya2021training,zhang2023knowledge,xu2023constructing,hong2023lasnn,guo2023joint,yu2025efficient}
to enhance the performance of student SNNs by pre-training an ANN as a guide for SNN training. Methods like \cite{takuya2021training,tran2022training} distill knowledge into an ANN teacher model, then convert the distilled ANN into an SNN. Other approaches, such as \cite{guo2023joint,hong2023lasnn}, follow a similar strategy by optimizing SNNs through a joint ANN framework. In addition, \cite{qiu2024self,zhangsupersnn} work to establish a mapping between the intermediate feature spaces of SNNs and ANNs, enabling direct feature distillation. However, refining the feature space for ANN-guided SNN learning continues to pose challenges that warrant further exploration. 


\section{Method}
\label{sec:3}

\subsection{Preliminary}

Each neuron in an SNN has a membrane potential and a threshold. After receiving a series of inputs, when the membrane potential reaches the threshold, it fires a binary spike and resets the membrane potential. We consider the commonly used LIF neuron model, whose equations for the membrane potential are given by:

\begin{equation}
\begin{cases}
u^{l}_{t+1} = \lambda(u^{l}_{t} - V_{th}s^{l}_{t}) + W^{l}s^{l-1}_{t} + b^{l}, \\
s^{l}_{t+1} = H(u^{l}_{t+1} - V_{th}),
\end{cases}
\end{equation}

\noindent
where $u^{l}_{t}$ and $s^{l}_{t}$ represent the membrane potential and fired spikes of the $l$-th layer neurons at time $t$, respectively. Here, $\lambda$ is the membrane potential constant, $V_{th}$ is the membrane threshold, $W^{l}$ denotes the weights of the $l$-th layer neurons, $b^{l}$ is the bias, and $H$ is the step function. In gradient surrogate methods, this function is typically replaced by differentiable functions such as the sigmoid.


\subsection{ANN-guided SNN Training Objectives}

We first partition the ANN model into different blocks based on depth and feature map size, aligning with current mainstream architectures that stack blocks. Initially, we perform rate encoding on the inputs, defining the ANN model as $ A = \{A_0, A_1, \ldots, A_n\} $ and the SNN model as $ S = \{S_0, S_1, \ldots, S_n\} $, where $ n $ represents the number of blocks in each model.

We then define a series of ANN-SNN hybrid models as follows:
\begin{equation}
 M_k = \{ S_0, S_1, \ldots, S_k, C_k, A_{k+1}^{frozen}, \ldots, A_n^{frozen}\}
\end{equation}
where $ C_k $ is a learnable module that maps features from the SNN rate-based representation space to the ANN representation space, with $ 0 \leq k < n $. The trainable weights are limited to the SNN model $S$ and $ C_k $, while the weights of the full ANN model sequence \( A = \{A_0, A_1, \ldots, A_n\} \), including \( \{A_{k+1}^{frozen}, \ldots, A_n^{frozen}\} \), remain fixed and are not updated.

We present the definition of the learning loss function for ANN-guided SNN training. Let the outputs of the models be defined as follows: the output of the ANN is denoted as $y_A $, the output of the SNN is represented as $y_S $, and the outputs of the hybrid models are given by $\{y_{M_k}\}_{k=1}^{n_b} $, where $ n_b $ is the number of hybrid models. The true labels are denoted as $ y $. To facilitate the efficient transfer of knowledge from the ANN to the SNN, we define the loss of the $ k $-th hybrid model as follows:
\begin{equation}
\label{loss1}
L_{blk}^{k} = (1 - \alpha) \cdot L_{ce}(y_{M_k}, y) + \alpha \cdot L_{kd}(y_{M_k}, y_A),
\end{equation}
where $ L_{ce} $ represents the cross-entropy loss for classification tasks, $ L_{kd} $ denotes the logit-based knowledge distillation loss defined via KL divergence, and $ \alpha $ is a hyperparameter that balances the contributions of the cross-entropy and distillation losses. The learning objective of our proposed ANN-SNN distillation framework is defined as:
\begin{equation}
\label{loss2}
L_{total} = (1 - \alpha) \cdot L_{ce}(y_S, y) + \alpha \cdot L_{kd}(y_S, y_A) + \sum_{k=1}^{n_b} \beta_k L_{blk}^{k}.
\end{equation}
Here, $\beta_k$  represents the loss weight for the hybrid model sequence $M_k$. Our detailed ANN-guided SNN distillation framework is illustrated in \cref{fig:main}.

\subsection{Implicit Feature Alignments}
We now outline the process through which we developed the approach for rate-based ANN-guided SNN distillation learning. 
The theoretical foundation of ANN2SNN is that the firing rate of neurons has an equivalent relationship with the ReLU activation function: $\phi^l_[T] = W^l \phi^{l-1}[T] - \frac{u^l[T] - u^l[0]}{T},$ where $\phi^l[T]$ represents the firing rate of the $l$-th layer SNN neurons after $T$ time steps, and $u^l[T]$ and $u^l[0]$ denote the membrane potential of the $l$-th layer neurons at time step $T$ and its initial value, respectively.

Let the output of the $l$-th layer neurons in the ANN be denoted as $a^{l}$, and the conversion error of the $l$-th layer can be defined as:
\begin{equation}
    \label{eq6}
    Err^{l} = \|\phi^{l}[T] - a^{l}\|.
\end{equation}

Inspired by the conversion error in ANN2SNN, we aim to define an ANN-guided loss in a similar form. This approach circumvents the challenge of explicitly defining the intermediate feature space alignment between ANN and SNN. The rate-based intermediate feature outputs of the model sequence $\{ S_0, S_1, \ldots, S_k\}$ are denoted as $F_{S_k}^{rate}$, while the intermediate feature outputs of the ANN model sequence $\{A_0, A_1, \ldots, A_k\}$ are denoted as $F_{A_k}$. A mapping $g_k$ is assumed to exist, which maps the rate-based feature $F_{S_k}^{rate}$ to $F_{A_k}$, expressed as $F_{S_k} = g_k\left(F_{S_k}^{rate}\right)$. This allows the distance between the SNN rate-based representation intermediate feature space and the ANN intermediate feature space to be formulated as follows, similar to \cref{eq6}: \begin{equation} 
Err^{k} = \|F_{S_k} - F_{A_k}\|. 
\end{equation}

Rather than enforcing strict imitation of the ANN’s intermediate features, the SNN is guided to achieve high task performance through effective learning. No explicit learning objective is set to minimize $Err^{k}$. Instead, $F_{S_k}$ is forwarded through the remaining ANN model sequence, and the loss is computed at the final logit layer, where the conversion error is inherently integrated into the loss calculation. For the knowledge distillation component, the loss is defined based on Kullback-Leibler (KL) divergence without temperature: \begin{equation} 
L_{kd}(x,y) = \text{KL}(y \parallel x) = \sum_{i} y_i \cdot \log \frac{y_i}{x_i},
\end{equation} where $ y_i $ represents the probability of class $ i $ as predicted by the teacher model, and $ x_i $ denotes the corresponding probability predicted by the student model.

Let $M^{p}$ denote the sequence after softmax operation on $\{A_{k+1}^{frozen},...,A_{n}^{frozen}\}$. $L_{kd}(y_{M_k}, y_A)$ is denoted as $L_{SA}$, and the distillation loss is formulated as:
\begin{equation}
\begin{aligned}
L_{SA} &= \text{KL}(M^{p}(F_{A_{k}}) || M^{p}(F_{S_{k}})) \\
       &= \sum_i M^{p}(F_{A_{k}})_i \log \frac{M^{p}(F_{A_{k}})_i}{M^{p}(F_{S_{k}})_i}.
\end{aligned}
\end{equation}
The gradient with respect to $F_{S_{k}}$ is derived as:
\begin{equation}
\label{eq10}
\frac{\partial L_{SA}}{\partial F_{S_{k}}} = \sum_i -\frac{M^{p}(F_{A_{k}})_i}{M^{p}(F_{S_{k}})_i} \cdot \frac{\partial M^{p}(F_{S_{k}})_i}{\partial F_{S_{k}}}.
\end{equation}
When the ANN output shows high confidence for class $i$ while SNN features processed through $M^p$ yield low confidence for the same class, the gradient is amplified, indicating strong ANN guidance for SNN learning. Conversely, if the ANN exhibits low confidence for class $i$ but $M^p$ outputs a high probability, the gradient is reduced, and updates are largely governed by the standard cross-entropy loss $L_{ce}(y_{M_k}, y)$. This approach prevents the SNN from over-adjusting to uncertain ANN predictions, thereby enhancing stability in the learning process.

Moreover, the training objective implicitly integrates learning that focuses on minimizing $Err^{k}$. We can approximate $M^p(F_{A_k})_i$ using first-order Taylor expansion:
\begin{equation}
\label{eq11}
 M^p(F_{A_k})_i \approx M^p(F_{S_k})_i + \frac{\partial M^p(F_{A_k})_i}{\partial F_{S_k}} \cdot (F_{A_k} - F_{S_k}).
\end{equation}
The gradient can be derived as follows:
\begin{equation}
\label{eq12}
\frac{\partial L_{SA}}{\partial F_{S_k}} =  \sum_i \left( \xi - 1  \right) \cdot \frac{\partial M^{p}(F_{S_{k}})_i}{\partial F_{S_{k}}},
\end{equation}
where:
\begin{equation}
    \xi = \frac{\partial M^p(F_{A_k})_i}{\partial F_{S_k}} \cdot \frac{F_{S_k} - F_{A_k}}{M^p(F_{S_k})_i}.
\end{equation}
\noindent
It can be observed that $Err^{k}$ is included in $\xi$. In \cref{4.4}, we demonstrate through experiments that our method facilitates the alignment of $F_{S_k}$  and $F_{A_k}$ , thereby reducing $Err^{k}$. This training objective enables the SNN to adaptively learn rate-based intermediate features, aligning them more effectively with the feature representations of the ANN model.

\begin{algorithm}[t]
\caption{ANN-guided SNN Distillation Training}
\label{alg:snn_ann_distillation}
\begin{algorithmic}[1]
    \REQUIRE 
        \STATE Pretrained ANN model $A = \{A_0, A_1, \ldots, A_n\}$
        \STATE SNN model $S = \{S_0, S_1, \ldots, S_n\}$
        \STATE Dataset $\mathcal{D}$
        \STATE Statistical parameters $\{\rho_{t}^l, g_t^l\}$ for rate-based backpropagation
    \ENSURE Trained SNN model $S$
    
\STATE Define $\{M_k^p = \{C_k, A_{k+1}^{\text{frozen}}, \ldots, A_n^{\text{frozen}}\}\}_{k=1}^{n_b}$

    \FOR{$x_{\text{batch}} \in \mathcal{D}$}
        \FOR{$l = 1$ to $L$}
            \FOR{$t = 1$ to $T$} 
                \STATE Update statistics $\{\rho_{t}^l, g_t^l\}$
            \ENDFOR
        \ENDFOR

        \STATE $y_A \leftarrow A^{frozen}(x_{\text{batch}})$ \COMMENT{Forward pass on ANN}
        \STATE $y_S, \{F_{S_k}\}_{k=1}^{n_b} \leftarrow S(x_{\text{batch}})$ \COMMENT{Forward pass on SNN}

        \FOR{$k = 1$ to $n_b$}
            \STATE $y_{M_k} \leftarrow M_k^p(F_{S_k})$ \COMMENT{Forward pass on $M_k^p$}
            \STATE Compute block loss $L_{blk}^k$ as in \cref{loss1}
        \ENDFOR

        \STATE Compute total loss $L_{total}$ as in \cref{loss2}
        \STATE Update SNN model parameters using $L_{total}$
    \ENDFOR
\end{algorithmic}
\end{algorithm}

\begin{table*}[t]
    \centering
    \small
    \setlength{\tabcolsep}{5pt}  
    \renewcommand{\arraystretch}{1.2}  
    \caption{\textbf{Performance comparison across CIFAR-10, CIFAR-100, and ImageNet datasets.} Results are averaged over five runs, except for ImageNet single-crop evaluations. $^*$ indicates that the SNN model architecture is SEW-ResNet-34. $^\dagger$ indicates that the ANN teacher model is ResNet-50.}
    \vspace{-10pt}
    \label{tab:main}
    \begin{tabular}{l l c c c c c c c}
        \toprule
        Method & Model & Time Steps & \multicolumn{2}{c}{CIFAR-10} & \multicolumn{2}{c}{CIFAR-100} & \multicolumn{2}{c}{ImageNet} \\
        & & & Top-1 Acc. (\%) &  & Top-1 Acc. (\%) &  & Top-1 Acc. (\%) &  \\
        \midrule
        STBP-tdBN~\cite{zheng2021going} & ResNet-19 & 4 & 92.92 & & - & & - & \\
        & ResNet-34 & 6 & - & & - & & 63.72 & \\
        Dspike~\cite{li2021differentiable} & ResNet-18 & 6 & 94.25 & & 74.24 & & - & \\
         &  & 4 & 93.66 & & 73.35 & & - & \\
         &  & 2 & 93.13 & & 71.68 & & - & \\
          & ResNet-34 & 6 & - & & - & & 68.19 & \\
        RecDis-SNN~\cite{guo2022recdis} & ResNet-19 & 6 & 95.55 & & - & & - & \\
        & & 4 & 95.53 & & 74.10 & & - & \\
        & & 2 & 93.64 & & - & & - & \\      
        TET~\cite{deng2022temporal} & ResNet-19 & 4 & 94.44 & & 74.47 & & - & \\
        & ResNet-34 & 6 & - & & - & & 64.79 & \\
        OTTT~\cite{xiao2022online} & ResNet-34 & 6 & - & & - & & 65.15 & \\
        SEW ResNet~\cite{fang2021deep} & ResNet-34$^*$ & 4 & - & & - & & 67.04 & \\
        Real Spike~\cite{guo2022real}& ResNet-34 & 4 & - & & - & & 67.69 & \\
        GLIF~\cite{yao2022glif}  & ResNet-18 & 6 & 94.88 & & 77.28 & & - & \\
        &  & 4 & 94.67 & & 76.42 & & - & \\
        &  & 2 & 94.15 & & 74.60 & & - & \\
        & ResNet-19 & 2 & 94.44 & & 75.48 & & - & \\
        & ResNet-34 & 4 & - & & - & & 67.52 & \\
        RateBP~\cite{yu2024advancing} & ResNet-18 & 6 & 95.90 & & 79.02 & & - & \\
        & & 4 & 95.61 & & 78.26 & & - & \\
        & & 2 & 94.75 & & 75.97 & & - & \\
        &ResNet-34 & 4 & - & & - & & 70.01 & \\
        \midrule

        LaSNN~\cite{hong2023lasnn} & VGG16  & 100  & 91.22 & & 61.52 & & - & \\
        KDSNN~\cite{xu2023constructing} & ResNet-18 & 4 & 93.41 & & - & & - & \\
        Joint A-SNN~\cite{guo2023joint} & ResNet-18 & 4 & 95.45 & & 77.39 & & - & \\
        & & 2 & 94.01 & & 75.79 & & - & \\
        EnOF~\cite{guoenof} & ResNet-19 & 2 & 96.19 & & 82.43 & & - & \\
        & ResNet-34 & 4 & - & & - & & 67.40 & \\
        TSSD~\cite{zuo2024self} & ResNet-18 & 2 & 93.37 & & 73.40 & & - & \\
        SAKD~\cite{qiu2024self} & ResNet-19 & 4 & 96.06 & & 80.10 & & - & \\
        & ResNet-34 & 4 & - & & - & & 70.04 & \\
        SuperSNN~\cite{zhangsupersnn} & ResNet-19 & 6 & 95.61 & & 77.45 & & - & \\
        \textbf{Ours} & ResNet-18 & 6 & \textbf{96.14} {\scriptsize $\pm$ 0.03} (96.17) & & \textbf{79.40} {\scriptsize $\pm$ 0.16} (79.50) & & - & \\
        & & 4 & \textbf{95.92} {\scriptsize $\pm$ 0.09} (96.01) & & \textbf{78.85} {\scriptsize $\pm$ 0.19} (79.08) & & - & \\
        & & 2 & \textbf{95.19} {\scriptsize $\pm$ 0.10} (95.31) & & \textbf{77.06} {\scriptsize $\pm$ 0.16} (77.25) & & - & \\
        & ResNet-19 & 2 & \textbf{96.56} {\scriptsize $\pm$ 0.06} (96.66) & & \textbf{81.44} {\scriptsize $\pm$ 0.12} (81.57) & & - & \\
        & ResNet-34*& 4 & - & & - & & \textbf{68.12} & \\
        & ResNet-34& 4 & - & & - & & \textbf{70.64} & \\
        & ResNet-34$^\dagger$& 4 & - & & - & & \textbf{71.76} & \\
        \bottomrule
    \end{tabular}
    \vspace{-5pt}
\end{table*}


\subsection{Rate-based backpropagation}

During SNN forward propagation, $T$ spike-based forward passes are performed without gradient tracking, with statistics required for backpropagation updated iteratively. This is followed by a single rate-based pass, at which point the gradient computation graph is established.
As shown in \cref{alg:snn_ann_distillation}, during the hybrid model $M_k$’s forward propagation, we operate only on the rate-based pass, with $C_k$ directly receiving rate-based SNN intermediate features. This approach enables the SNN to learn rate-based representations efficiently.

Specifically, define the average rate as $ r^l = \mathbb{E}[s^l_t] = \frac{1}{T} \sum_{t \leq T} s^l_t $. The input to the linear layer is given by $ I^l_t = W^l s^{l-1}_t $, and the average rate of the input can be expressed as $ c^l = \mathbb{E}[I^l_t] = \mathbb{E}[W^l s^{l-1}_t] = W^l \mathbb{E}[s^{l-1}_t] = W^l r^{l-1} $. We use $ c^l $ to estimate $ I^l_t $, and the weight gradient is formulated as follows:

\begin{equation}
(\nabla_w L)_\text{rate} = \frac{\partial L}{\partial c^l} \cdot \frac{\partial c^l}{\partial W^l} = \frac{\partial L}{\partial c^l} \cdot {r^{l-1}}^\top.
\end{equation}

According to \cite{yu2024advancing}, an equivalent form for $ \frac{\partial L}{\partial c^l} $ is derived:
\begin{equation}
\frac{\partial L}{\partial c^l} = \left( \frac{\partial L}{\partial c^L} \prod_{i=L-1}^l \left( W^{i\top} \mathbb{E}\left[\frac{\partial s_t^l}{\partial u_t^l} \rho_t^l\right] \right) \right).
\end{equation}
Here, the statistical term is defined as:
\begin{equation}
\rho_t^l = 1 + \rho_{t-1}^l \left( \frac{\partial u_t^l}{\partial u_{t-1}^l} + \frac{\partial u_t^l}{\partial s_{t-1}^l} \frac{\partial s_{t-1}^l}{\partial u_{t-1}^l} \right).
\end{equation}
Furthermore, define $ g_t^l $ as follows:
\begin{equation}
g_t^l = \frac{1}{t} \left( (t-1) g_{t-1}^l + \frac{\partial s_t^l}{\partial u_t^l} \rho_t \right).
\end{equation}
The gradients contained in the statistics $ \rho_t^l $ and $ g_t^l $ can be computed directly. As demonstrated in \cite{yu2024advancing}, the expected value $\mathbb{E} \left[ \frac{\partial s_t^l}{\partial u_t^l} \rho_t^l \right] $ can be obtained as:
\begin{equation}
g_T^l = \mathbb{E} \left[ \frac{\partial s_t^l}{\partial u_t^l} \rho_t^l \right]. 
\end{equation}
This allows us to derive \( \frac{\partial L}{\partial c^l} \) and subsequently perform a single backpropagation to update the weights.

\subsection{ANN-guided SNN Distillation Framework}

Our entire ANN-guided SNN Distillation Framework is outlined in \cref{alg:snn_ann_distillation}. Initially, we define the ANN components of the $ n_b $ hybrid models, denoted as $ \{M_k^p\}_{k=1}^{n_b} $. For each mini-batch of data, the ANN performs a forward pass to generate the teacher labels $ y_A $. The SNN then executes $ T $ forward passes without recording gradients to update the statistical parameters for backpropagation, $ \{\rho_t^l, g_t^l\} $. This is followed by a single forward pass that records gradients to obtain the student labels $ y_S $ and the intermediate rate features $ \{F_{S_k}\}_{k=1}^{n_b} $. 

Next, we conduct the forward pass for each $ M_k^p $ to obtain the corresponding outputs $ y_{M_k} $. We first compute the loss for each hybrid model $ L_{blk}^k $ and then aggregate these to determine the total loss $ L_{total} $. Finally, we perform backpropagation of the loss, wherein the gradients for the SNN parameters are computed using rate-based backpropagation, leveraging $ \{\rho_T^l, g_T^l\} $ to decouple the time dimension during SNN training.

It is worth noting that our approach, in terms of implementation, requires only minimal modifications during the forward propagation phase, and there are no structural constraints on the teacher and student models.

\section{Experiments}
\label{sec:4}

In this section, we evaluate the effectiveness of the proposed ANN-guided SNN distillation framework, focusing on the accuracy at different time steps for image classification tasks. \cref{4.2} present a comparative analysis of our approach against existing SNN direct training methods and several established SNN distillation techniques. \cref{4.3} provides a comprehensive overview of the ablation studies, while \cref{4.4} analyzes the effectiveness of our proposed approach in conjunction with experimental results.

\begin{figure}[t]
\centering
\includegraphics[width=1.0\columnwidth]{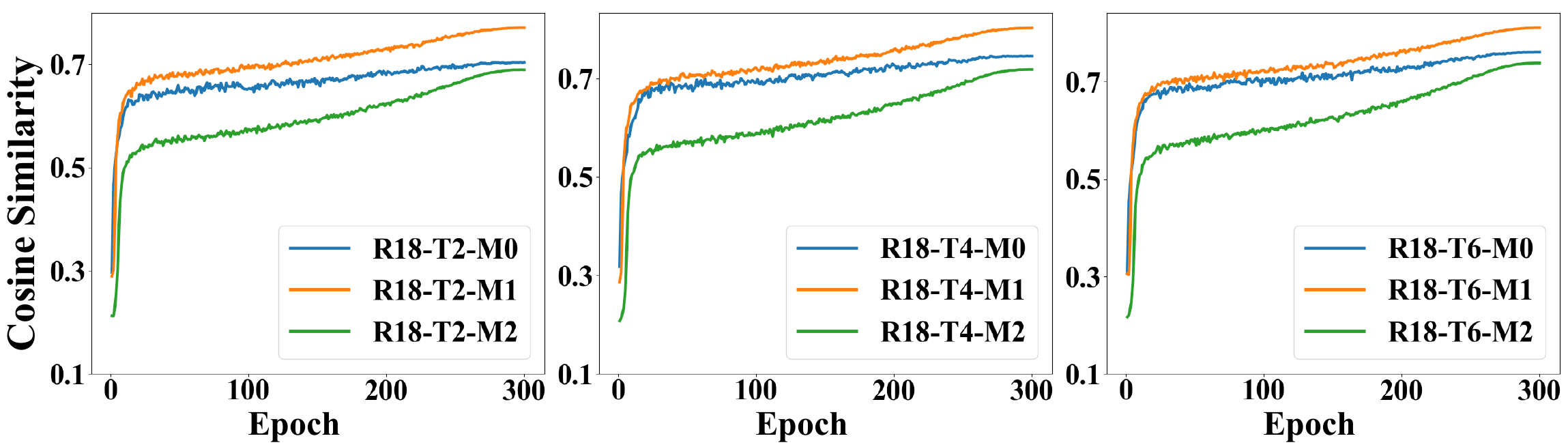} 
\vspace{-5pt}
\caption{\textbf{Measures of feature similarity.} The results of cosine similarity distances are obtained by ResNet-18 on CIFAR-100. Each subplot is labeled according to the naming convention “R18(ResNet-18)-T(timesteps)-M($M_k$).”}
\label{Cosine_sim}
\end{figure}

\begin{figure}[t]
\centering
\includegraphics[width=1.0\columnwidth]{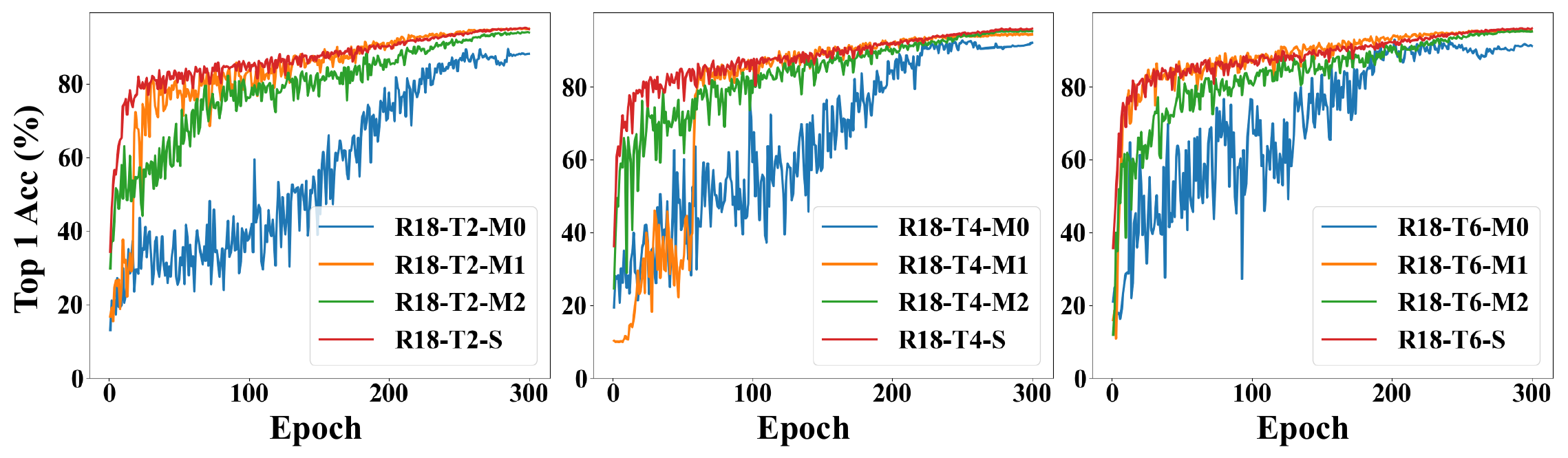} 
\vspace{-5pt}
\caption{\textbf{Validation accuracy of ANN-SNN hybrid models and SNN model during training.} The results are obtained by ResNet-18 on CIFAR-10. Each subplot is labeled according to the naming convention “R18(ResNet-18)-T(timesteps)-M($M_k$)/S(SNN).”}
\label{Block_acc}
\end{figure}

\begin{figure}[t]
\centering
\includegraphics[width=1.0\columnwidth]{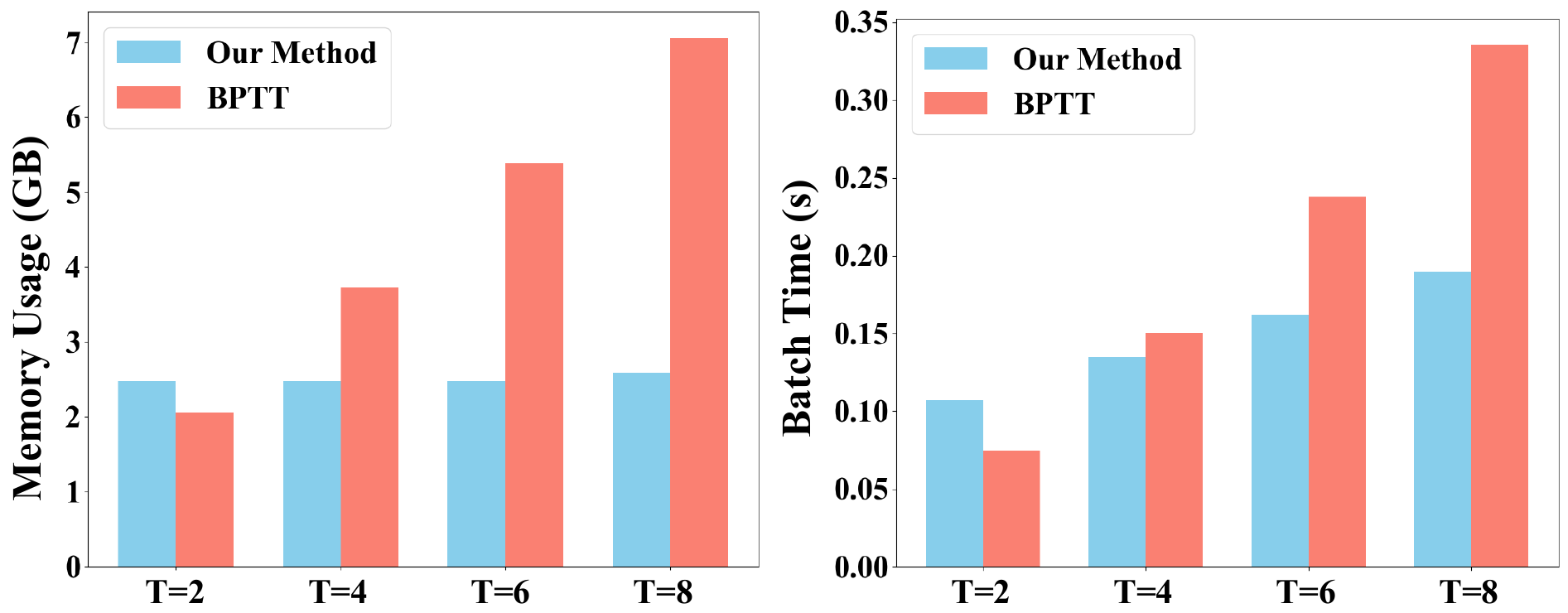} 
\vspace{-5pt}
\caption{\textbf{Comparison of Training overhead between our method and BPTT for direct training.} The results are obtained by averaging over three epochs of stable operation with ResNet-18 on CIFAR-100, using a single NVIDIA 3090 GPU.}
\label{Training_overhead}
\end{figure}

\subsection{Experimental Setup}
\noindent
\textbf{Dataset.}
We conduct experiments on three static image classification datasets and one neuromorphic dataset: ImageNet \cite{deng2009imagenet}, CIFAR-10, CIFAR-100 \cite{krizhevsky2009learning}, and CIFAR10-DVS \cite{li2017cifar10}. ImageNet comprises ~1.2 million training and 50,000 validation images across 1,000 classes. CIFAR-10 includes 60,000 images of size 32x32 across 10 categories, while CIFAR-100 offers a more challenging task with 60,000 images across 100 categories. CIFAR10-DVS, based on dynamic vision sensors (DVS), contains event streams adapted from CIFAR-10 images, making it suitable for neuromorphic model evaluation. 

\noindent
\textbf{Training Setting.}
Our experiments use models from the ResNet family. For CIFAR and CIFAR10-DVS, models are trained for 300 epochs with a learning rate of 0.1, weight decay of 5e-4, and a batch size of 128. For ImageNet, we use 100 epochs with a 0.2 learning rate, 2e-5 weight decay, and batch size 512. More detailed training specifics are provided in the supplement.

\subsection{Comparison to the State-of-the-Art}
\label{4.2}

\noindent
\textbf{Results on CIFAR.}
We use ResNet-18 and ResNet-19 as student models. For ResNet-18, the teacher model is an ANN of the same structure, achieving accuracies of 96.92\% and 79.95\% on CIFAR-10 and CIFAR-100, respectively. For ResNet-19, the teacher model is ResNet-34 due to their structural similarity, achieving accuracies of 97.24\% and 81.90\%. Like ASGL\cite{wang2023adaptive}, we apply AutoAugment \cite{cubuk2018autoaugment} and Cutout \cite{devries2017improved} for data augmentation. \cref{tab:main} displays the performance on CIFAR-10 and CIFAR-100 datasets. Compared to other SNN methods based on BPTT and existing SNN distillation techniques, our approach achieves or surpasses state-of-the-art results. For ResNet-18, our method shows significant improvements over Joint A-SNN \cite{guo2023joint}, KDSNN \cite{xu2023constructing} and TSSD~\cite{zuo2024self} across all time steps. For ResNet-19, at $T=2$, we surpass the accuracy of SAKD \cite{qiu2024self} at $T=4$ and TSSD~\cite{zuo2024self} at $T=6$.

\noindent
\textbf{Results on ImageNet.}
We employ SEW-ResNet-34 \cite{fang2021deep} and the commonly used PreAct-ResNet-34 \cite{hu2021advancing} as student models, with the teacher model also being ResNet-34, achieving an accuracy of 76.32\%. Additionally, we conduct experiments using ResNet-50 as the teacher model, where the student model, PreAct-ResNet-34, achieved an accuracy of 71.76\%. Our experimental results, presented in \cref{tab:main}, demonstrate that our method outperforms existing SNN approaches based on BPTT and previously established SNN distillation methods, achieving state-of-the-art performance. Experiments on static image datasets indicate that our approach effectively transfers knowledge from the ANN pretrained models to the SNN.

\noindent
\textbf{Results on CIFAR10-DVS.}
We use the ResNet-19 as both the student and teacher models on the neuromorphic dataset CIFAR10-DVS. The ANN teacher model is implemented by replacing SNN neurons with ReLU, and its input is the mean over the time dimension, achieving an accuracy of 83.60\%. Similar to previous work \cite{li2021differentiable}, we split the dataset into training and testing images, with the training set containing 9,000 images and the testing set containing 1,000 images. For data preprocessing and augmentation, we first resize the training images to 48×48 pixels and then randomly flip them horizontally. Additionally, we apply a random rolling operation within a range of 5 pixels. The testing images are directly resized to 48×48 pixels. \cref{tab:dvs} shows the performance on CIFAR10-DVS, where our method achieves state of the art with shorter time steps.

\subsection{Ablation Study}
\label{4.3}

\noindent
\textbf{Ablation study on the position of $M_k$.}
To identify optimal insertion points for ANN-guided feature learning, we perform ablation experiments with ResNet-18 on CIFAR-100 (see \cref{tab:ablation}). Vanilla logit-based knowledge distillation provides only a 0.15\% accuracy improvement over direct training, while adding ANN-guided branches further enhances performance. For ResNet-18, we segment the model into four main layers, with a head section preceding them and a classification layer following. We denote the hybrid models inserted after the head as $M_H$, after each main layer (Layers 1–3) as $M_0$ through $M_2$, and before the classification layer as $M_{fc}$.
Results show that positioning ANN-guided branches at the outputs of main layers achieves the best performance. Omitting any branch in these layers degrades accuracy, underscoring the value of feature alignment across layers. This may stem from distributional differences between ANN and SNN knowledge, as intermediate SNN outputs within main layers appear to align more effectively with ANN feature spaces, thereby enhancing the impact of ANN-guided learning.

\noindent
\textbf{Control experiment of hyperparameters $\alpha$ and $\beta_k$.} 
We examine different strategies for $\alpha$ and $\beta_k$ in \cref{tab:control}. For $\alpha$, three options are tested: an "increase" strategy that linearly interpolates from 0 to 1 over epochs, a "decrease" strategy with the opposite trend, and a fixed value of 0.5, which balances the ANN-guided loss and hard-label loss. For $\beta_k$, we evaluate decay by powers of 2 in descending order of $k$ and a uniform assignment of $1/n_b$, where $n_b$ is the number of hybrid models. Results suggest that setting $\alpha = 0.5$ and $\beta_k = 1/n_b$ yields optimal performance.

\begin{table}[tb]
    \centering

    \small  
    \renewcommand{\arraystretch}{1.0}  
    \caption{\textbf{Performance on CIFAR10-DVS.} Results are averaged over five runs of experiments. }
    \vspace{-5pt}
    \label{tab:dvs}
    \resizebox{1\linewidth}{!}{
    \begin{tabular}{cccc}
    \toprule[1.5pt]
    Method & Model & Timesteps & Top-1 Acc (\%) \\
     
    \midrule[1.3pt]
    \multirow{1}{*}{Rollout} \cite{kugele2020efficient} & DenseNet& 10&66.80 \\
    \hline
    \multirow{1}{*}{LIAF-Net} \cite{wu2021liaf} & LIAF-Net& 10&71.70 \\
    \hline
      \multirow{1}{*}{STBP-tdBN \cite{zheng2021going}} & \multirow{1}{*}{ResNet-19} 
    & 10 & 67.80 \\
    \hline
       \multirow{1}{*}{RecDis-SNN \cite{guo2022recdis}} & \multirow{1}{*}{ResNet19} 
    & 10 & 72.42 \\
    \hline
       \multirow{1}{*}{Real Spike \cite{guo2022real}} & \multirow{1}{*}{ResNet-19} 
    & 10 & 72.85 \\
    \hline
       \multirow{1}{*}{Dspike \cite{li2021differentiable}} & \multirow{1}{*}{ResNet-18 } 
    & 10 & 75.4 \\
    \hline
       \multirow{1}{*}{GLIF \cite{yao2022glif}} & \multirow{1}{*}{7B-wideNet} 
    & 16 & 78.10 \\  
    \hline
       \multirow{1}{*}{RateBP \cite{yu2024advancing}} & \multirow{1}{*}{VGG-11} & 10 & 76.96 \\
    \hline
      ENOF~\cite{guoenof} & ResNet19& 10 & 80.10 \\
     \hline
       SAKD \cite{qiu2024self} & ResNet-19 & 4 & 80.30\\
    \hline
    Ours & ResNet-19 & 4 & \textbf{80.54}\textpm0.71(81.70) \\
\bottomrule[1.5pt]
    \end{tabular}
    }
\end{table}

\begin{table}[]
    \centering
    \caption{\textbf{Ablation study of the design of the position of $M_k$.} The results are obtained using ResNet-18 with $T = 4$ on CIFAR-100, averaged over five runs of experiments.}
    \vspace{-5pt}
    \begin{tabular}{c|ccccc|c}
    \toprule
         $y_A$ &\multicolumn{1}{c}{$y_{M_H}$} &\multicolumn{1}{c}{$y_{M_0}$} &$y_{M_1}$
          &$y_{M_2}$
          & $y_{M_{fc}}$
          &\multicolumn{1}{c}{Accuracy (\%)} \\
    \hline
     &&&&&& 78.26\textpm0.16\\
        \ding{51} &&&&&& 78.41\textpm0.27\\
        \ding{51} &\ding{51}&\ding{51}&\ding{51}&\ding{51}&\ding{51}& 78.69\textpm0.07\\
        \ding{51} &&\ding{51}&\ding{51}&\ding{51}&\ding{51}& 78.70\textpm0.06\\
        \ding{51} &&\ding{51}&\ding{51}&\ding{51}&& \textbf{78.85}\textpm0.19\\
      \ding{51} &&&\ding{51}&\ding{51}&& 78.75\textpm0.10\\
      \ding{51} &&\ding{51}&&\ding{51}&& 78.72\textpm0.15\\
      \ding{51} &&\ding{51}&\ding{51}&&& 78.68\textpm0.17\\
    \bottomrule
    \end{tabular}
    \label{tab:ablation}
\end{table}


\begin{table}[tb]
    \centering
    \caption{\textbf{Control experiment of hyper-parameter $\alpha$ and $\beta_k$. } $\uparrow$ represents "increase", and $\downarrow$ represents "decrease".}
    \vspace{-5pt}
     \renewcommand{\arraystretch}{1.5}
    \resizebox{1\linewidth}{!}{
    \begin{tabular}{c|c|c|c|c|c|c}
    \toprule[1.5pt]
    $\beta_k$ & \multicolumn{3}{c|}{$1/2^{n-k}$} & \multicolumn{3}{c}{$1/n_b$} \\
    \hline
    $\alpha$ & $\uparrow$ & $\downarrow$ & 0.5 & $\uparrow$ & $\downarrow$ & 0.5  \\
    \hline
    Top-1 Acc(\%) & 78.34{\scriptsize $\pm$ 0.11}   & 78.21{\scriptsize $\pm$ 0.21} & 78.29{\scriptsize $\pm$ 0.11} & 78.28{\scriptsize $\pm$ 0.16} &  78.52{\scriptsize $\pm$ 0.20} & \textbf{78.85}{\scriptsize $\pm$ 0.19}\\
    \bottomrule[1.5pt]
    \end{tabular}
    \label{tab:control}
    }
\end{table}

\subsection{Analysis and Discussion}
\label{4.4}

\noindent
\textbf{Cosine similarity between intermediate features.} 
Cosine similarity measures the alignment between two vectors by quantifying the cosine of the angle between them, with values approaching 1 indicating closer directional alignment. As shown in \cref{Cosine_sim}, we use this metric to examine the correspondence between the SNN’s rate-based intermediate feature  space ($F_{S_k}$) and the ANN’s feature space ($F_{A_k}$) throughout ResNet-18 training at time steps $T=2$, $4$, and $6$. Although no explicit objective is set for minimizing $ \|F_{S_k} - F_{A_k}\| $, cosine similarity between $F_{S_k}$ and $F_{A_k}$ increases consistently across training. Among the hybrid models, $M_1$ shows the highest similarity at all time steps, followed by $M_0$, with $M_2$ showing the lowest. This pattern suggests that intermediate feature mappings in dense backbone layers are more aligned with those of the ANN, while layers nearer the classifier exhibit a task-focused orientation.

\noindent
\textbf{Validation accuracy of $M_k$ during training.} 
In the process of ANN-guided SNN training, we record the validation accuracy trends of ResNet-18 and the ANN-SNN hybrid models on CIFAR-10 across different time steps, as shown in \cref{Block_acc}. The validation accuracy trends for $M_k$ closely resemble the cosine similarity patterns observed in \cref{Cosine_sim}. Specifically, the validation accuracies of $M_1$ and $M_2$ closely align with that of the SNN, while $M_0 $ shows slightly lower accuracy. We interpret the hybrid models $M_k$ as forming a series of unique teacher models, establishing a self-distillation-like framework that effectively guides the learning of the SNN.

\noindent
\textbf{Training overhead comparing to BPTT.} 
We compare the training cost of ANN-guided SNN distillation with that of BPTT-based direct training, as shown in \cref{Training_overhead}. At $T=2$, our method shows slightly higher batch time and memory usage than BPTT; however, as the time steps increase, our method does not exhibit a linear increase in memory usage and batch time like BPTT. At $T=4, 6,$ and $8$, both memory and computation time are significantly lower than BPTT. This efficiency is due in part to rate-based backpropagation, as well as the relatively low number of additional parameters introduced by our method—such as only 0.17M additional parameters for $C_k$ in ResNet-18 and ResNet-34. These results highlight the training efficiency of our approach. 

\vspace{-8pt}
\section{Conclusion}
\label{sec:5}
\vspace{-5pt}
In this paper, we proposed a rate-based ANN-guided SNN distillation framework that incorporates pretrained ANN models via a block-wise replacement approach. This method allows for flexible alignment of SNN feature spaces with rate-based representations, achieving effective knowledge transfer. Our experiments demonstrate that the proposed framework achieves state-of-the-art performance, and we hope it provides valuable insights that contribute to further advancements in ANN-guided SNN learning.


\vspace{-8pt}
\section*{Acknowledgments}
\vspace{-8pt}
This work was supported by Zhejiang Provincial Natural Science Foundation of China under Grant No. ZCLZ24F0102, the National Natural Science Foundation of China under Grant No. 62304203, 62401501, 62431012 and 62027805, the National Key Research and Development Program of China under Grant No. 2023YFB3812500.

{
    \small
    \bibliographystyle{ieeenat_fullname}
    \bibliography{main}
}



\end{document}